\documentclass[10pt,twocolumn,letterpaper]{article}
\PassOptionsToPackage{table}{xcolor}
\usepackage[pagenumbers]{cvpr}
\usepackage[T1]{fontenc}
\newcommand{\sd}[1]{{\scriptsize$\pm$#1}}
\newcommand{\ours}{FLUENT}
\ifdefined\XeTeXversion
  \IfFileExists{msam10.pfb}{}{%
    \renewcommand{\checkmark}{\text{\begingroup
      \setlength{\unitlength}{0.1em}%
      \begin{picture}(9,8)
        \put(0,3){\rotatebox{-45}{\rule{0.37em}{0.055em}}}%
        \put(2.6,0.4){\rotatebox{53}{\rule{0.90em}{0.055em}}}%
      \end{picture}\endgroup}}}
\fi
\renewcommand{\eqref}[1]{(\ref{#1})}
\definecolor{cvprblue}{rgb}{0.21,0.49,0.74}
\usepackage[pagebackref,breaklinks,colorlinks,allcolors=cvprblue]{hyperref}
\hypersetup{
  hypertexnames=false,
  pdftitle={Practice Makes Fluent: Learning to Use Earlier Predictions in Looped Models},
  pdfauthor={Minseok Seo, Hee-Seon Kim, Changick Kim}
}
\title{Practice Makes Fluent: Learning to Use Earlier Predictions in Looped Models}
\author{%
  \href{https://openreview.net/profile?id=~Minseok_Seo1}{\textcolor{black}{Minseok Seo}}\ensuremath{^{\dagger}}
  \qquad
  \href{https://openreview.net/profile?id=~Hee-Seon_Kim1}{\textcolor{black}{Hee-Seon Kim}}\ensuremath{^{\dagger}}
  \qquad
  \href{https://openreview.net/profile?id=~Changick_Kim1}{\textcolor{black}{Changick Kim}}\\[3pt]
  {\normalsize School of Electrical Engineering, Korea Advanced Institute of Science and Technology (KAIST)}\\
  {\small
    \href{mailto:minseok.seo@kaist.ac.kr}{minseok.seo@kaist.ac.kr}\quad
    \href{mailto:hskim98@kaist.ac.kr}{hskim98@kaist.ac.kr}\quad
    \href{mailto:changick@kaist.ac.kr}{changick@kaist.ac.kr}}\\[2pt]
  {\small\ensuremath{^{\dagger}}\,These authors contributed equally.}%
}
\begin{document}

\twocolumn[{
  \renewcommand\twocolumn[1][]{#1}%
  \maketitle
  \begin{minipage}{\textwidth}
  \centering
  \includegraphics[width=\linewidth]{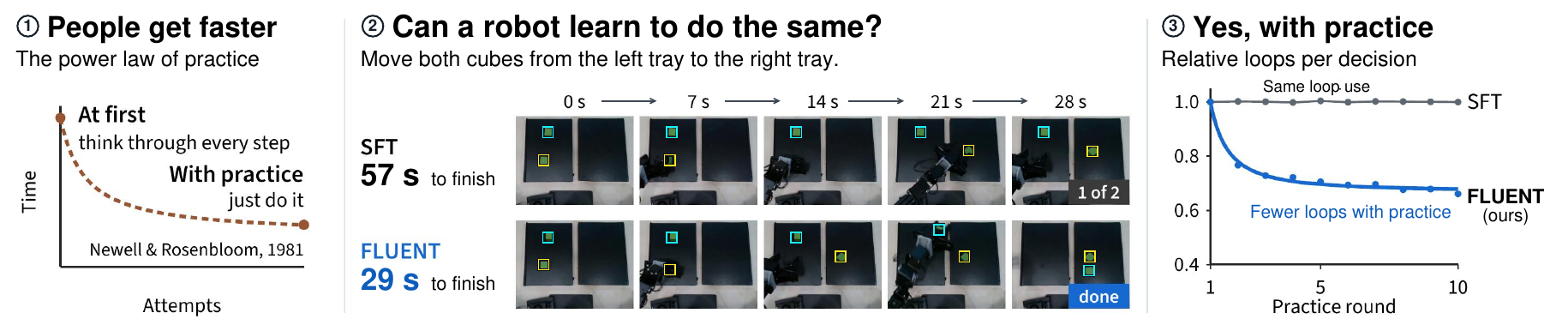}
  \captionof{figure}{\textbf{Practice changes which predictions a looped policy can use.} The center shows end-to-end real-robot pick-and-place: successful episodes take 31\,s on average for \ours{} and 69\,s for SFT (illustrated: 29\,s and 57\,s), with fewer chunks and lower per-chunk time. }
  \label{fig:practice_overview}
  \end{minipage}\par\medskip
}]

\begin{abstract}
People become faster at a task through practice, but models that improve with experience do not necessarily learn to use less computation.
We train looped models, whose intermediate predictions allow earlier exits, to perform tasks with fewer iterations through practice.
This is difficult because shared-weight updates can degrade the final-loop prediction while improving early ones, and the exit gate must track early predictions as their quality changes.
We propose \ours{}, which requires no additional task labels or rewards: it distills the initial model's final-loop prediction into early predictions, anchors the current final-loop prediction to the same frozen teacher, and trains the exit gate to balance disagreement with the current final-loop prediction against computation cost.
\ours{} improves performance at shallower depth in new Crafter worlds, raises LIBERO success with fewer action-head iterations, and maintains language-model accuracy at lower selected depth.
On a real robot, success rises from 62.3\% to 74.6\% while mean time over successful episodes falls from 69\,s to 31\,s. The project page is available at \url{https://msseok96-cpu.github.io/FLUENT/}.

\end{abstract}

\begin{figure*}[t]
\centering
\includegraphics[width=\linewidth]{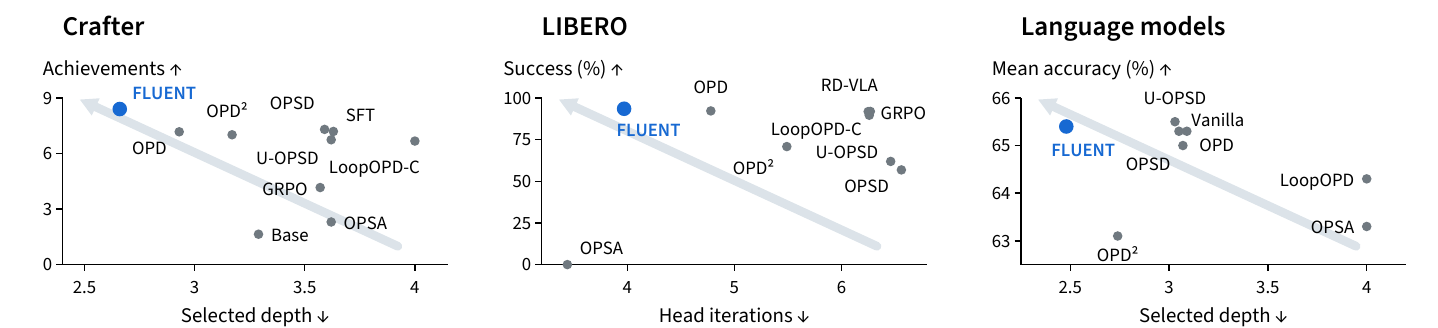}
\vspace{-6mm}
\caption{\textbf{Performance against depth in three benchmark settings.} Points use Tables~\ref{tab:crafter_main}, \ref{tab:libero_main}, and~\ref{tab:ouro_suite}; upper left is better. Crafter and language models use selected output depth, whereas LIBERO uses executed action-head iterations. Arrows indicate the preferred direction.
}
\label{fig:tradeoffs}
\vspace{-3mm}
\end{figure*}

\section{Introduction}

A person who repeats a task gets faster at it. Execution time often falls
steeply over the first attempts and more gradually thereafter~\cite{newell2013mechanisms,logan1992shapes}
(Fig.~\ref{fig:practice_overview}, left).
Robots and agents also perform similar tasks repeatedly, from
visual manipulation~\cite{brohan2023rt,kim2024openvla,black2024pi_0,tur2026recurrent}
to language-guided interaction~\cite{shinn2023reflexion,wang2023voyager}.
Yet better task performance through experience does not necessarily
mean less computation at inference.

Early exit and adaptive-depth methods allocate computation according
to the current input~\cite{schuster2022confident,xin2020deebert,liu2020fastbert,zhu2025scaling,yue2024deer,tur2026recurrent}.
Our question is complementary: \textbf{can experience make a previously
difficult task require less computation?}
Methods that reuse reasoning traces~\cite{ouyang2026reasoningbank,shinn2023reflexion,chen2026log},
skills~\cite{wang2023voyager,ouyang2026skillos}, or cached
computation~\cite{gim2024prompt,yao2025cacheblend,chun2026functional}
exploit previous experience through external records.
These records must be stored, retrieved, and matched to the current task.
We instead store experience in the model's weights: the model
learns through self-distillation from the task inputs it encounters,
without additional labels or rewards. We call this process \emph{practice},
and aim to improve task performance while reducing the computation needed
to perform the task.

We study this question in looped models~\cite{zhu2025scaling,jeddi2026loopformer,tur2026recurrent}.
In the architectures we consider, a shared block repeatedly updates
a latent state, and a shared output head can read a prediction
after each loop.
This makes earlier predictions available, but availability alone
does not make them useful.
Practice must both improve these predictions and enable the policy
to act on them.
A natural starting point is to use the frozen initial model as a
teacher, distilling its deep predictions into shallow loops.

However, this addresses only part of this problem.
It directly supervises agreement between the first loop and the
teacher, not the task performance of every subsequent loop.
Because the weights are shared, later predictions also change,
but they need not improve together.
Moreover, even an improved first-loop prediction saves computation
only when the exit gate selects it.
Improving early predictions and learning to use them are
distinct aspects of practice.

We present \ours{}, a self-distillation procedure that addresses
these two parts together.
Translate trains early predictions to match the final-loop output
of the frozen initial model.
Because these updates also affect later loops,
Anchor keeps the current final-loop prediction close to the
same frozen teacher.
This anchored prediction provides the reference for training the gate.
Gate adapts the exit policy to the updated predictions, balancing
agreement with the current final loop against the cost of further
computation.
Practice uses no additional task labels or rewards, and inference
requires no external experience store.

We evaluate \ours{} in four settings, examining both prediction
depth and task performance.
In Crafter~\cite{hafner2021benchmarking}, \ours{} improves performance
in new worlds while selecting shallower predictions.
On LIBERO~\cite{liu2023libero}, \ours{} executes 36.7\% fewer
action-head iterations than RD-VLA~\cite{tur2026recurrent},
while improving success.
On a real robot, relative to the SFT policy, \ours{} increases
success from 62.3\% to 74.6\% and reduces mean completion time
over successful episodes from 69\,s to 31\,s.
In language-model distillation, compared with the vanilla model
using its original gate, \ours{} maintains mean accuracy
while reducing mean selected depth by 18.7\%.

\begin{figure*}[t]
\centering
\includegraphics[width=0.9\linewidth]{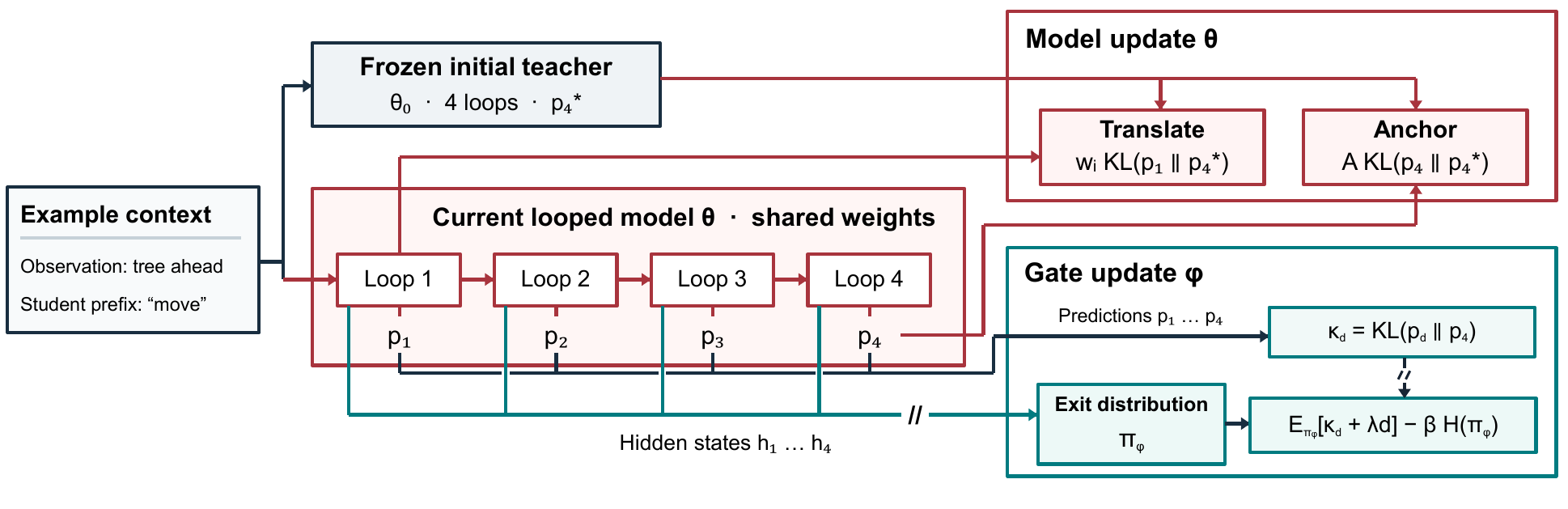}
\caption{\textbf{Overview of \ours{}.} The initial model's final prediction supervises early predictions (Translate) and the student's final prediction (Anchor). The gate learns a trade-off between disagreement with the current final prediction and depth (Gate). Deep discrepancies are training targets, not inference-time inputs to the gate. Double slashes denote stopped gradients.}
\label{fig:practice_method}
\end{figure*}

\section{Related Work}
\label{rel}

\noindent\textbf{Looped models and adaptive depth.}
Adaptive inference allocates computation across inputs or exits from
intermediate predictions~\citep{damani2410learning,snell2024scaling,schuster2022confident,xin2020deebert,liu2020fastbert}.
Ouro learns predictions and an exit distribution across recurrent
depths~\citep{zhu2025scaling}; LoopFormer aligns short and long
paths~\citep{jeddi2026loopformer}. In manipulation, RD-VLA refines an action
head recurrently~\citep{tur2026recurrent}, while other methods use intermediate
actions or learned stopping~\citep{yue2024deer,izzo2026decoupled,lei2026think}.
Policy distillation can also reduce a fixed inference depth~\citep{jeon2026shallow}.
We study changes in both predictions and exits during practice.
Tokenwise early termination requires a cache policy such as Ragged KV~\citep{mishra2026enabling}.

\vspace{1mm}
\noindent\textbf{Reusing experience.}
Experience-dependent reasoning cost has been studied with memory-based
systems~\citep{pan2025speedupllm}. Agents store traces, feedback, and executable
skills for later reuse~\citep{ouyang2026reasoningbank,ouyang2026skillos,yang2024buffer,wang2024agent,shinn2023reflexion,wang2023voyager}.
LAG represents past reasoning logs as KV states, and FCGraft reuses functions
and their caches to accelerate code-policy
generation~\citep{chen2026log,chun2026functional}. Other cache systems reduce
repeated prompt processing~\citep{gim2024prompt,zheng2024sglang,yao2025cacheblend}.
Our setting updates policy parameters rather than retrieving prior episodes
or cached task solutions at inference; it is complementary to retrieval.

\vspace{1mm}
\noindent\textbf{On-policy and self-distillation.}
On-policy distillation trains on contexts generated by the
student~\citep{agarwal2024policy,heo2026policy}. OPSD and U-OPSD construct
supervision from the model's own predictions under different
contexts~\citep{zhao2026self,li2026policy}; OPSA instead adapts using
entropy-derived advantages without a teacher~\citep{ding2026does}.
These methods do not explicitly optimize recurrent exit depth, even when they
improve other efficiency measures. Self-distilled multi-exit networks also
teach intermediate predictions~\citep{liu2020fastbert,zhang2019byot,phuong2019distillation},
and distillation can preserve a prior model's outputs during further
learning~\citep{li2017lwf}. We use these principles jointly in a recurrent
policy whose shallow and deep predictions share the same block.

\vspace{1mm}
\noindent\textbf{Distillation across loops.}
LoopOPD distills a frozen four-loop teacher into a two-loop student,
and D-LoopOPD uses the current four-loop prediction as the
teacher~\citep{wang2026recurrent}; both report gains at intermediate
and final depths but compare methods on four-loop outputs and learn
no exit policy.
Such gains are not guaranteed under shared-weight updates, motivating our explicit preservation of the final-loop prediction.
LoOPSD distills a teacher built from searched
layer-repetition routes into a standard-path
student~\citep{fan2026loopsd}. A better shallow prediction yields an
earlier output only if the exit policy selects it, and shared-weight
updates shift the final prediction the gate is trained against.
\ours{} therefore pairs Anchor, which keeps the final prediction
aligned with the frozen initial teacher, with Gate, which balances
prediction discrepancy and computation cost.

\section{Method}
\label{sec:method}

\subsection{Preliminaries: looped models and early exit}
\label{sec:prelim}
For context $c$, a shared block with parameters $\theta$ produces
predictions $p_d^\theta(\cdot\mid c)$ at depths $d=1,\ldots,D$.
We restrict attention to architectures with a depth-wise readout
through a shared output module.
The token gate has parameters $\phi$ and stopping hazards based
on the hidden states available up to each depth. These hazards define an exit distribution $\pi_\phi(d\mid c)$.
For threshold $q\in(0,1]$, we select
\begin{equation}
e(c)=\min\Big\{d:
\textstyle\sum_{j=1}^{d}\pi_\phi(j\mid c)\ge q
\Big\}.
\label{eq:exit_rule}
\end{equation}
Lower $e(c)$ means an earlier output.

\subsection{Practice: translating deep predictions to a shallow loop}
\label{sec:practice}
\label{sec:translate}
Let $\theta_0$ denote the model parameters before practice.
We initialize the student at $\theta_0$ and keep the initial model
frozen, using its final-loop prediction as the teacher:
$p_D^\star(\cdot\mid c)=p_D^{\theta_0}(\cdot\mid c)$.
In interactive settings, practice proceeds in rounds: the agent interacts with the environment using its current policy and then trains on the states it visited.

On each visited state, the shallow student ($d=1$) generates
a response. For each token context $c_1,\dots,c_N$ in that response,
Translate matches the shallow prediction to the frozen teacher:
\begin{equation}
\mathcal L_{\mathrm T}
=
\frac{1}{N}\sum_{i=1}^{N} w_i\,
\mathrm{KL}\!\left(
p_1^\theta(\cdot\mid c_i)
\,\middle\|\,
p_D^\star(\cdot\mid c_i)
\right).
\label{eq:translate}
\end{equation}
The weight $w_i$ is proportional to the highest token probability
of the teacher, placing more weight on contexts where the teacher
is confident.
Because the student generates the response itself, distillation
uses token contexts encountered by the shallow student rather
than teacher-generated contexts.
The frozen model supervises practice, so no additional rewards, labels, or demonstrations are required.

\subsection{Anchor: holding the final loop in place}
\label{sec:anchor}
Shared parameters couple the shallow and final predictions.
An update to the shallow prediction can therefore change
the final output, even with a frozen teacher. We constrain this change by aligning the current final prediction
with the same frozen teacher:
\begin{equation}
\begin{aligned}
\mathcal L_{\mathrm A}
&=
\frac{1}{N}\sum_{i=1}^{N}
\mathrm{KL}\!\left(
p_D^\theta(\cdot\mid c_i)
\,\middle\|\,
p_D^\star(\cdot\mid c_i)
\right),
\\
\mathcal L_{\mathrm{LM}}
&=
\mathcal L_{\mathrm T}+A\mathcal L_{\mathrm A}.
\end{aligned}
\label{eq:model_loss}
\end{equation}
Here $A>0$ controls the strength of the anchor.
Anchor constrains predictions on practice contexts;

\begin{table*}[t!]
\centering
\caption{\textbf{Crafter performance and selected depth.} Seen and Unseen denote training and new worlds. Results average seven seeded repetitions; $\pm$ denotes their standard deviation.
Loops averages per-game mean selected token depth over Seen and Unseen games; readout is dense except in the ragged rows.
 Bold marks the largest displayed performance mean or smallest depth in each column.}
\label{tab:crafter_main}
\renewcommand{\arraystretch}{1.08}
\setlength{\tabcolsep}{3.93897pt}
\small
\begin{tabular}{l c ccccc  ccccc}
\toprule
 & & \multicolumn{5}{c}{Ouro-1.4B} & \multicolumn{5}{c}{Ouro-2.6B} \\
\cmidrule(lr){3-7}\cmidrule(lr){8-12}
 & & \multicolumn{2}{c}{Seen} & \multicolumn{2}{c}{Unseen} & & \multicolumn{2}{c}{Seen} & \multicolumn{2}{c}{Unseen} & \\
\cmidrule(lr){3-4}\cmidrule(lr){5-6}\cmidrule(lr){8-9}\cmidrule(lr){10-11}
Method & Exit & Ach.$\uparrow$ & Score$\uparrow$ & Ach.$\uparrow$ & Score$\uparrow$ & Loops$\downarrow$ & Ach.$\uparrow$ & Score$\uparrow$ & Ach.$\uparrow$ & Score$\uparrow$ & Loops$\downarrow$ \\
\midrule
Base & Gate & 1.00\sd{0.00} & 0.4 & 1.64\sd{0.05} & 0.7 & 3.29 & 2.04\sd{0.33} & 1.4 & 1.89\sd{0.32} & 1.3 & 3.30 \\
SFT & Gate & 10.81\sd{0.38} & 22.7 & 7.19\sd{0.58} & 12.6 & 3.63 & 12.61\sd{0.16} & 26.0 & 9.97\sd{0.85} & 25.3 & 3.40 \\
\midrule
GRPO~\citep{shao2024deepseekmath} & Gate & 5.01\sd{1.82} & 9.5 & 4.16\sd{1.16} & 6.7 & 3.57 & 6.45\sd{1.32} & 14.2 & 5.50\sd{1.88} & 10.6 & 3.36 \\
OPD~\citep{agarwal2024policy} & Gate & 8.83\sd{0.62} & 15.7 & 7.17\sd{0.68} & 11.9 & 2.93 & \textbf{12.76}\sd{0.53} & 29.7 & 10.67\sd{0.60} & 24.2 & 2.90 \\
OPD$^2$~\citep{heo2026policy} & Gate & 8.11\sd{0.36} & 14.7 & 7.01\sd{0.70} & 10.7 & 3.17 & 11.93\sd{0.40} & 26.7 & 9.70\sd{1.32} & 22.3 & 3.15 \\
OPSA~\citep{ding2026does} & Gate & 2.82\sd{0.87} & 5.5 & 2.30\sd{0.85} & 2.8 & 3.62 & 3.44\sd{0.61} & 7.7 & 2.93\sd{0.70} & 4.4 & 3.42 \\
OPSD~\citep{zhao2026self} & Gate & 8.69\sd{0.74} & 17.2 & 7.30\sd{0.61} & 10.2 & 3.59 & 8.99\sd{0.61} & 20.5 & 7.14\sd{0.97} & 13.8 & 3.40 \\
U-OPSD~\citep{li2026policy} & Gate & 7.83\sd{0.63} & 13.8 & 6.74\sd{0.58} & 10.5 & 3.62 & 12.11\sd{0.99} & 29.1 & 9.60\sd{1.13} & 20.7 & 3.39 \\
LoopOPD~\citep{wang2026recurrent} & Fixed-4 & 9.08\sd{0.32} & 18.8 & 6.67\sd{0.91} & 11.3 & 4.00 & 12.49\sd{0.33} & 29.0 & 9.51\sd{0.74} & 23.4 & 4.00 \\
\midrule
\rowcolor{blue!5}
\textbf{\ours{}} & Gate & \textbf{11.54}\sd{0.50} & \textbf{26.1} & \textbf{8.40}\sd{0.94} & \textbf{16.5} & 2.66 & 12.69\sd{0.40} & \textbf{30.8} & \textbf{11.31}\sd{1.11} & \textbf{27.0} & 2.58 \\

\rowcolor{blue!5}
\textbf{\ours{}} (ragged) & Gate & 9.25\sd{1.00} & 25.5 & 8.02\sd{0.28} & 15.1 & \textbf{2.52} & 12.29\sd{0.41} & 29.5 & 10.30\sd{0.47} & 26.7 & \textbf{2.50} \\
\bottomrule
\end{tabular}
\end{table*}

\subsection{Gate: learning when to stop}
\label{sec:gate}
As practice changes the predictions at different depths, the exit policy must adapt to determine which predictions to use. We train the gate to balance agreement with the current final prediction against the cost of additional computation. For token context $c_i$, the prediction cost at depth $d$ is
\begin{equation}
\kappa_{i,d}
=
\mathrm{KL}\!\left(
p_d^\theta(\cdot\mid c_i)
\,\middle\|\,
p_D^\theta(\cdot\mid c_i)
\right).
\label{eq:gate_cost}
\end{equation}
The gate minimizes
\begin{equation}
\begin{aligned}
\mathcal L_{\mathrm G}
=
\frac{1}{|\mathcal M|}
\sum_{i\in\mathcal M}\Big[
&\sum_{d=1}^{D}
\pi_\phi(d\mid c_i)
(\kappa_{i,d}+\lambda d)
\\
&-\beta H(\pi_\phi(\cdot\mid c_i))
\Big].
\end{aligned}
\label{eq:gate_loss}
\end{equation}
Here $\mathcal M$ is the set of response-token positions included in gate training, $\lambda$ controls the depth penalty, and $\beta$ weights the entropy $H$ of the exit distribution.
The prediction cost favors agreement with the current final loop,
while the depth penalty favors earlier exits when agreement is comparable.
The entropy term encourages a less concentrated exit distribution
during training.
The gate reference is the current final prediction $p_D^\theta$,
whereas Translate and Anchor use the frozen teacher $p_D^\star$.
Anchor connects these two references by limiting how far the current final prediction moves from the teacher.
At inference, the learned gate selects the output depth
using Eq.~\eqref{eq:exit_rule}; costs
$\kappa_{i,d}$ are training signals.

\vspace{1mm}
\noindent\textbf{Overall training.}
Combining the model objective in Eq.~\eqref{eq:model_loss}
with the gate objective in Eq.~\eqref{eq:gate_loss}, we write
\begin{equation}
\mathcal L_{\mathrm{total}}
=
\mathcal L_{\mathrm T}
+
A\mathcal L_{\mathrm A}
+
\mathcal L_{\mathrm G}
\bigl(\phi;\operatorname{sg}[\theta]\bigr),
\label{eq:total_loss}
\end{equation}
where $\operatorname{sg}$ denotes stop-gradient.
The hidden states and prediction costs used for gate training
are detached from the model.
Thus, $\theta$ is updated using
$\mathcal L_{\mathrm T}+A\mathcal L_{\mathrm A}$,
whereas $\phi$ is updated using $\mathcal L_{\mathrm G}$.
Gate gradients do not modify the shared model.
Figure~\ref{fig:practice_method} summarizes the full pipeline:
the frozen deep teacher supervises shallow predictions and
anchors the current final prediction, while the gate learns
when to use the updated predictions.

\begin{table*}[t!]
\centering
\caption{\textbf{LIBERO success and executed action-head iterations.}
Left: each method's default stopping rule (consecutive-action MSE for baselines, learned gate for \ours{}).
Right: four fixed iterations.
Evaluation uses 30 new initial states per task and seven repetitions
(8,400 episodes per method per exit).
$\pm$ is the standard deviation of the seven overall rates.
Bold/underline mark first/second displayed values.
$^\dagger$Uniform supervision includes the final iteration but no extra Anchor penalty.}
\vspace{-3mm}
\label{tab:libero_main}
\renewcommand{\arraystretch}{1.08}
\small
\setlength{\tabcolsep}{7.60482pt}
\begin{tabular}{l cccccc  ccccc}
\toprule
& \multicolumn{6}{c}{Default stopping rule}
& \multicolumn{5}{c}{Success at four iterations (\%)} \\
\cmidrule(lr){2-7}\cmidrule(lr){8-12}
Method
& Spatial & Object & Goal & Long & Avg. & Iters$\downarrow$
& Spatial & Object & Goal & Long & Avg. \\
\midrule
RD-VLA~\citep{tur2026recurrent}
& \underline{96.7} & \underline{96.8} & 93.6
& \underline{80.4} & 91.9\sd{0.58} & 6.27
& 88.1 & 92.4 & 85.8 & 62.3 & 82.2\sd{0.71} \\
\midrule
GRPO~\citep{shao2024deepseekmath}
& 96.9 & 96.4 & 93.6 & 80.4 & 91.8\sd{0.44} & 6.25
& 89.2 & 93.1 & 87.3 & 65.0 & 83.7\sd{0.66} \\
OPD~\citep{agarwal2024policy}
& \textbf{97.1} & 96.7 & \textbf{95.4}
& 79.4 & \underline{92.2}\sd{0.51} & 4.78
& \underline{96.6} & \underline{97.1} & \textbf{95.4}
& \underline{77.8} & \underline{91.7}\sd{0.60} \\
OPD$^2$~\citep{heo2026policy}
& 83.3 & 75.1 & 81.9 & 43.0 & 70.8\sd{1.92} & 5.49
& 77.3 & 76.4 & 84.0 & 38.6 & 69.1\sd{2.10} \\
OPSA~\citep{ding2026does}
& 0.0 & 0.0 & 0.0 & 0.0 & 0.0\sd{0.00} & \textbf{3.44}
& 0.0 & 0.0 & 0.0 & 0.0 & 0.0\sd{0.00} \\
OPSD~\citep{zhao2026self}
& 77.7 & 95.2 & 9.2 & 45.1 & 56.8\sd{2.41} & 6.56
& 66.7 & 94.2 & 5.4 & 22.9 & 47.3\sd{2.77} \\
U-OPSD~\citep{li2026policy}
& 69.9 & 92.1 & 21.2 & 64.6 & 61.9\sd{2.18} & 6.46
& 51.6 & 82.7 & 14.0 & 43.3 & 47.9\sd{2.63} \\
LoopOPD~\citep{wang2026recurrent}
& 93.4 & 95.2 & 91.3 & 78.6 & 89.6\sd{0.72} & 6.26
& 85.1 & 90.1 & 88.9 & 70.0 & 83.5\sd{0.95} \\
\midrule
\rowcolor{blue!5}
\textbf{\ours{}}
& 96.1 & \textbf{99.0} & \underline{95.0}
& \textbf{83.9} & \textbf{93.5}\sd{0.62} & \underline{3.97}
& \textbf{97.3} & \textbf{97.2} & \underline{94.5}
& \textbf{80.2} & \textbf{92.3}\sd{0.41} \\
\bottomrule
\end{tabular}
\vspace{-3mm}
\end{table*}

\begin{table}[t!]
\centering
\caption{\textbf{Real-robot pick-and-place.} Success is over all attempted episodes; Time and Chunks average successful episodes. Iters is mean executed head iterations per chunk. $\pm$ denotes standard deviation across seven repetitions. Episodes use a 300\,s timeout.}
\vspace{-3mm}
\label{tab:robot_main}
\footnotesize
\setlength{\tabcolsep}{3.49943pt}
\begin{tabular}{lccccc}
\toprule
Method & Succ.\ (\%)$\uparrow$ & Time (s)$\downarrow$ & Chunks & s/chunk & Iters$\downarrow$ \\
\midrule
RD-VLA (SFT) & 62.29\sd{1.60} & 69 & 86 & 0.80 & 8.25 \\
OPD & 70.43\sd{1.90} & 42 & 61 & 0.69 & 5.50 \\
w/o Anchor & 66.14\sd{2.27} & 36 & 58 & 0.62 & 3.90 \\
\midrule
SFT, 32 iterations & 65.71\sd{1.98} & 125 & 70 & 1.78 & 32 \\
SFT, 4 iterations & 47.86\sd{2.41} & 60 & 97 & 0.62 & 4 \\
\ours{}, delayed & 69.86\sd{1.86} & 49 & 61 & 0.80 & \textbf{3.46} \\
\midrule
\rowcolor{blue!5}
\textbf{\ours{}} & \textbf{74.57}\sd{1.72} & \textbf{31} & 52 & 0.60 & \textbf{3.46} \\
\bottomrule
\end{tabular}
\vspace{-3mm}
\end{table}
\begin{table*}[t]
\centering
\caption{\textbf{Ouro-1.4B language evaluation.} Avg is the unweighted mean of twelve accuracy metrics. Loops is the equal-weight mean of generated-token selected depths on MMLU-Pro, BBH, GSM8K, and MATH500.
Evaluation uses dense readout. $\pm$ denotes standard deviation of the aggregate over seven repetitions; no standard deviation was reported for the fixed pretrained Vanilla reference.}
\label{tab:ouro_suite}
\renewcommand{\arraystretch}{1.08}
\setlength{\tabcolsep}{3.2pt}
\resizebox{\textwidth}{!}{%
\begin{tabular}{@{}l c *{14}{c}@{}}
\toprule
 & & \multicolumn{6}{c}{General} & \multicolumn{2}{c}{Math} & \multicolumn{2}{c}{HumanEval} & \multicolumn{2}{c}{MBPP} & \multicolumn{2}{c}{Summary} \\
\cmidrule(lr){3-8}\cmidrule(lr){9-10}\cmidrule(lr){11-12}\cmidrule(lr){13-14}\cmidrule(l){15-16}
Method & Exit & MMLU & MMLU-Pro & BBH & ARC-C & HellaSwag & WinoGrande & GSM8K & MATH500 & Base & Plus & Base & Plus & Avg & Loops$\downarrow$ \\
\midrule
Vanilla & Gate & \textbf{68.3} & 41.1 & \textbf{64.9} & 59.1 & \textbf{72.4} & 69.1 & \textbf{69.9} & 72.4 & 70.1 & 63.4 & 72.2 & \textbf{60.8} & 65.3 & 3.05 \\
\midrule
OPD~\citep{agarwal2024policy} & Gate & 67.9 & \textbf{41.2} & 64.3 & 58.8 & 72.2 & 69.1 & 68.2 & 73.4 & \textbf{71.3} & \textbf{65.2} & 72.5 & 60.1 & 65.3\sd{0.2} & 3.09 \\
OPD$^2$~\citep{heo2026policy} & Gate & 66.8 & 40.4 & 61.7 & 57.6 & 71.5 & 68.2 & 63.2 & 74.0 & 65.2 & 59.8 & 69.8 & 59.0 & 63.1\sd{0.5} & 2.74 \\
OPSA~\citep{ding2026does} & Fixed-4 & 67.4 & 38.9 & 63.6 & \textbf{59.4} & 71.4 & 68.0 & 66.9 & 70.6 & 65.9 & 60.4 & 69.6 & 57.1 & 63.3\sd{0.4} & 4.00 \\
OPSD~\citep{zhao2026self} & Gate & 68.0 & 40.5 & 64.4 & 58.7 & 72.1 & 69.0 & 69.1 & 73.0 & 69.5 & 63.4 & 72.0 & 59.9 & 65.0\sd{0.3} & 3.07 \\
U-OPSD~\citep{li2026policy} & Gate & 68.2 & 40.2 & 64.8 & 58.9 & 72.3 & 69.5 & 69.8 & 74.0 & 70.1 & 64.6 & \textbf{73.5} & 60.1 & \textbf{65.5}\sd{0.2} & 3.03 \\
LoopOPD~\citep{wang2026recurrent} & Fixed-4 & 67.1 & 40.6 & 63.7 & 59.2 & 71.9 & 68.7 & 68.8 & 72.8 & 67.1 & 62.8 & 70.4 & 58.2 & 64.3\sd{0.3} & 4.00 \\
\rowcolor{blue!5}
\textbf{\ours{}} & Gate & 67.9 & 40.7 & 63.5 & 59.3 & 72.2 & \textbf{69.9} & 68.8 & \textbf{74.4} & 70.7 & \textbf{65.2} & 71.7 & 60.6 & 65.4\sd{0.2} & \textbf{2.48} \\
\bottomrule
\end{tabular}}
\end{table*}

\section{Experiments}
\label{sec:exp}
\subsection{Setup}
\label{sec:setup}

\noindent\textbf{Models and what is trained.}
In Crafter~\cite{hafner2021benchmarking}, Ouro-1.4B and
Ouro-2.6B~\cite{zhu2025scaling} use up to four loops. We start from
demonstration-trained SFT and update the shared model and gate. In
LIBERO~\cite{liu2023libero}, we start from public
RD-VLA~\cite{tur2026recurrent}, freeze the vision--language and proprioception
modules, and train the recurrent action head and a new exit gate. On the real
robot, initialization uses five hours of teleoperation and 200 hours of
synthetic Isaac Sim data. The primary experiment reuses their input states, with the frozen
iteration-eight output as target, without additional robot rollouts. The language
experiment trains Ouro-1.4B for 100 steps on the math, science, and code prompt
mixture used by OPD$^2$~\cite{heo2026policy}.

\vspace{1mm}
\noindent\textbf{Continuous actions.}
Our framework also applies to models that produce continuous
actions rather than token distributions.
We replace KL divergence with MSE in the distillation, anchoring,
and gate objectives.
Translate transfers deep teacher predictions to shallow iterations,
Anchor constrains the prediction at the teacher depth, and Gate balances
prediction discrepancy against iteration cost.

\vspace{1mm}
\noindent\textbf{RD-VLA setting.}
For RD-VLA~\cite{tur2026recurrent}, the action head runs for up to
$D=32$ iterations, and the action can be read out after any of them.
The teacher is the output of the frozen initial model after eight
iterations. Translate distills it into the student's output at
iteration two, and Anchor holds the student's output at iteration
eight to the same target:
\begin{equation}
\mathcal L_{\mathrm{act}}=
\underbrace{
\mathrm{MSE}(a_2^\theta,a_8^\star)
}_{\text{Translate}}
+
\underbrace{
3\,\mathrm{MSE}(a_8^\theta,a_8^\star)
}_{\text{Anchor}}.
\label{eq:action_loss}
\end{equation}
Here $a_8^\star$ is the action output of the frozen initial
model after eight iterations.
The gate uses normalized MSE to the current final action output
as its prediction cost and $\gamma d/32$ as its iteration cost
in Eq.~\eqref{eq:gate_loss}.
The suite-specific normalization is calibrated once before
training and then fixed.

\vspace{1mm}
\noindent\textbf{Baselines.}
We compare adapted OPD, OPD$^2$, OPSA, OPSD, U-OPSD, and
LoopOPD~\cite{agarwal2024policy,heo2026policy,ding2026does,zhao2026self,li2026policy,wang2026recurrent}.
For a controlled comparison, we adapt these methods to use
deep-loop predictions as teacher targets and shallow-loop
predictions as student outputs, without introducing an external
teacher model.
Within each setting, all methods start from the same initial
checkpoint as \ours{}.

\subsection{Crafter: higher performance at lower depth}
\label{sec:experience_eval}
\label{sec:results}

\noindent\textbf{Benchmark and protocol.}
\label{sec:baselines}
\label{sec:metrics}
Crafter is a procedurally generated survival game with
22 distinct achievements~\cite{hafner2021benchmarking}.
We run ten practice rounds, each containing two games in fixed
training worlds and one game in a newly generated world.
Only states from the training worlds enter practice.
This setting evaluates both improvement in repeatedly encountered
worlds and transfer to new ones.
Table~\ref{tab:crafter_main} aggregates play before the update
in each round.
Seen and Unseen denote training and new worlds, respectively.
The main comparison uses dense readout.

\vspace{1mm}
\noindent\textbf{Results.}
\ours{} improves Seen and Unseen achievements over SFT at both
model sizes while selecting shallower predictions.
On Ouro-1.4B, Unseen achievements increase from 7.19 to 8.40
(16.8\%), while mean selected depth decreases from 3.63 to 2.66
(26.7\%).
Ouro-2.6B shows the same pattern:
Unseen achievements increase from 9.97 to 11.31 (13.4\%),
and mean selected depth decreases from 3.40 to 2.58 (24.1\%).

The achievement gains are especially pronounced in new worlds.
At both model sizes, the improvement over SFT is larger in
Unseen worlds than in Seen worlds.
Among the methods evaluated under dense readout, \ours{} achieves the highest
Unseen mean achievements and Score, together with the lowest
mean selected depth.
Methods such as OPD remain competitive in the practice worlds,
but \ours{} leads when new-world performance and selected
depth are considered together.

These results capture the fluency we seek from practice:
the benefits extend beyond the worlds already visited.
The practiced policy uses earlier predictions while performing
better in new worlds.

Using an earlier prediction does not by itself skip computation in Ouro. Later tokens attend to the key--value cache of earlier tokens at every loop, so under dense readout a token that exits early is still computed to the final loop. To test whether the practiced policy also works when computation is actually skipped, we run inference with Ragged KV~\citep{mishra2026enabling}, under which each generated token stops at its selected loop.
This additional execution setting reaches the lowest mean selected depth in \cref{tab:crafter_main}, with new-world achievements above SFT at both model sizes (8.02 vs.\ 7.19; 10.30 vs.\ 9.97).

\subsection{Real robot: faster and more reliable}
\label{sec:real_robot}

\noindent\textbf{Task.}
A 6-DoF arm moves specified objects to designated locations. The 100 scenarios vary the number (one to seven), shapes, and positions of objects, with one to three objects specified for relocation.
Evaluation layouts are excluded from training. An episode succeeds when all specified objects reach their target locations.
We repeat each scenario seven times. Completion time is measured on the robot from episode start to success and averaged over successful episodes. An episode that does not succeed within 300\,s is counted
as a failure.

\vspace{1mm}
\noindent\textbf{Results.}
\ours{} improves both task success and completion time. Success increases from 62.29\% for SFT to 74.57\%, while mean completion time over successful episodes decreases from 69\,s to 31\,s, a 2.23$\times$ ratio. Executed action-head iterations decrease by 58.1\%, from 8.25 to 3.46. Successful episodes also use fewer action chunks on average, from 86 to 52, while mean time per chunk decreases
from 0.80\,s to 0.60\,s.

The success gain persists when action timing is matched. Delaying \ours{} to the SFT cadence of 0.80\,s per chunk still yields 69.86\% success, compared with 62.29\% for SFT. Without the added delay, \ours{} reaches 74.57\%. Meanwhile, SFT with four fixed iterations has a similar per-chunk execution time but lower success.

\subsection{LIBERO: success with fewer iterations}
\label{sec:robot_eval}

\noindent\textbf{Benchmark and protocol.}
LIBERO comprises Spatial, Object, Goal, and Long
suites~\cite{liu2023libero}.
We distill teacher predictions at iteration eight into
student predictions at iteration two.
We train suite-specific action heads for three rounds
on recorded policy inputs.
At evaluation, \ours{} uses its learned gate to select
the number of executed iterations.
Each method is evaluated on 30 new initial states for each
of 40 tasks over seven repetitions, giving 8,400 episodes
per stopping rule.
Table~\ref{tab:libero_main} reports results under the default
stopping rule of each method and with four fixed iterations.

\vspace{1mm}
\noindent\textbf{Results.}
The initial RD-VLA policy already achieves approximately
92\% success.
Our primary focus is therefore how few iterations can
sustain high control performance, with additional success
gains reported as a secondary benefit.
\ours{} achieves 93.5\sd{0.62}\% success at 3.97 iterations
per chunk, compared with 91.9\sd{0.58}\% at 6.27 for RD-VLA.
It thus executes 36.7\% fewer action-head iterations
while achieving a higher mean success rate.

The fixed-four evaluation makes the improvement at
shallower depth particularly clear.
The original RD-VLA head achieves 82.2\% success,
whereas \ours{} reaches 92.3\% at the same iteration count:
a 10.1-percentage-point improvement.
\ours{} also achieves the highest fixed-four mean among
the compared methods, including OPD at 91.7\%.
Practice therefore does more than advance the stopping point:
it enables earlier iterations to produce actions that
support high task success.

\subsection{Language models: earlier outputs after fixed-corpus distillation}
\label{sec:reasoning_eval}

\noindent\textbf{Benchmark and protocol.}
This experiment tests whether useful predictions can become
available at shallower depth using a fixed prompt corpus,
without collecting new environment interactions.
We train Ouro-1.4B for 100 steps on an equal mixture of
math, science, and code prompts.
The deep prediction of the initial model provides the teacher,
and distillation uses response contexts generated by
the shallow student.

We evaluate general knowledge and reasoning, mathematics,
and code generation using MMLU, MMLU-Pro, BBH, ARC-C,
HellaSwag, WinoGrande, GSM8K, MATH500, HumanEval/+,
and MBPP/+~\cite{hendrycks2020measuring,wang2024mmlupro,suzgun2023bbh,clark2018arc,zellers2019hellaswag,sakaguchi2020winogrande,cobbe2021training,hendrycks2021math,lightman2024lets,chen2021evaluating,austin2021program,liu2023evalplus}.
Mean accuracy gives equal weight to the twelve evaluation
metrics.
Mean selected depth averages generated-token depth across
MMLU-Pro, BBH, GSM8K, and MATH500.
Evaluation uses dense readout.

\begin{figure*}[t]
\centering
\includegraphics[width=1.0\linewidth]{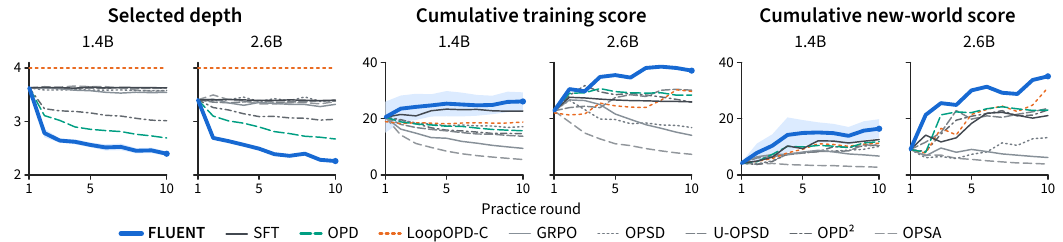}
\caption{\textbf{Depth and cumulative Crafter score over practice.} Curves show seven-run means at 1.4B (\ours{} shading: $\pm1$ SD) and a single-run illustration at 2.6B. Seven-repetition aggregate results at both sizes appear in Table~\ref{tab:crafter_main}. Selected depth and cumulative score have different aggregation rules.
}
\label{fig:crafter_rounds}
\end{figure*}

\vspace{1mm}
\noindent\textbf{Results.}
\ours{} maintains the mean accuracy of the initial vanilla
model while using shallower predictions
(Table~\ref{tab:ouro_suite}).
Mean accuracy is 65.4\%, compared with 65.3\% for
the vanilla model, while mean selected depth decreases
from 3.05 to 2.48, a reduction of 18.7\%.
Among the compared methods, \ours{} achieves the lowest
mean selected depth while retaining the original level
of mean accuracy.

The baseline comparison highlights the importance of
considering accuracy and selected depth together.
OPD reaches 65.3\% mean accuracy, but its selected depth
remains 3.09.
U-OPSD reaches 65.5\% at a selected depth of 3.03.
By comparison, \ours{} uses predictions at a substantially
shallower mean depth of 2.48 with comparable mean accuracy.
OPD$^2$ also reduces selected depth, to 2.74, but its mean
accuracy falls to 63.1\%.

These results extend \ours{} beyond interactive environments.
Even with a fixed set of prompts, cross-depth distillation
and exit-policy updates enable the model to use earlier
predictions while preserving mean accuracy.

\subsection{Ablation study}
\label{sec:ablation}

\noindent\textbf{Component ablation.}
Translate, Anchor, and Gate play complementary roles
(Table~\ref{tab:crafter_ablation}).
Translate alone reduces mean selected depth from 3.63
for SFT to 2.97, while unseen achievements change
from 7.19 to 7.00.
Adding Anchor raises mean achievements to 8.36.

Gate retraining makes the improved performance available
at shallower depth.
Adding Gate to Translate and Anchor reduces mean selected
depth from 3.24 to 2.66, with 8.40 mean achievements.
At the same rounded selected depth of 2.66, the no-Anchor
variant earns 7.05 achievements.
This comparison highlights the importance of combining
early exit with prediction preservation.

Gate-only retraining also helps, reaching 7.70 achievements
at 3.15 loops.
The full method combines the highest achievement mean
with the joint-lowest selected depth.
Together, shallow-prediction learning, deep-prediction
preservation, and exit-policy updates enable \ours{}
to achieve higher task performance at shallower depth.

\begin{table}[t!]
\centering
\caption{\textbf{Effects of Translate, Anchor, and Gate
in Crafter} (Ouro-1.4B).
Checkmarks indicate enabled training components;
Gate denotes exit-policy retraining.
Achievements and Score concern Unseen worlds;
Loops averages selected token depth over Seen and Unseen
games under dense readout.
Bold marks the best displayed means
(higher Ach.\ and Score, lower Loops).
}
\label{tab:crafter_ablation}
\footnotesize
\setlength{\tabcolsep}{8.15307pt}
\begin{tabular}{cccccc}
\toprule
\multicolumn{3}{c}{Training}
& \multicolumn{2}{c}{Unseen} & \\
\cmidrule(lr){1-3}\cmidrule(lr){4-5}
Translate & Anchor & Gate & Ach. & Score & Loops \\
\midrule
-- & -- & --
& 7.19\sd{0.58} & 12.6 & 3.63 \\
-- & -- & $\checkmark$
& 7.70\sd{0.46} & 14.7 & 3.15 \\
\midrule
$\checkmark$ & -- & --
& 7.00\sd{0.34} & 11.5 & 2.97 \\
$\checkmark$ & -- & $\checkmark$
& 7.05\sd{0.86} & 11.7 & \textbf{2.66} \\
$\checkmark$ & $\checkmark$ & --
& 8.36\sd{0.73} & \textbf{16.5} & 3.24 \\
\rowcolor{blue!5}
$\checkmark$ & $\checkmark$ & $\checkmark$
& \textbf{8.40}\sd{0.94}
& \textbf{16.5} & \textbf{2.66} \\
\bottomrule
\end{tabular}
\end{table}

\noindent\textbf{Change over practice.}
As practice progresses, \ours{} uses shallower predictions
while improving task performance (Fig.~\ref{fig:crafter_rounds}).
Selected depth decreases sharply during the first rounds
and more gradually thereafter, following the qualitative
pattern illustrated in Fig.~\ref{fig:practice_overview}.
Meanwhile, cumulative Score improves in both training
and new worlds.
Practice therefore makes the policy more proficient
with fewer loops, including in environments it has not practiced in.
Refreshing experience also helps: at equal update counts,
collecting new states each round yields a mean selected
depth of 2.66, compared with 2.89 when repeatedly training
on first-round states.

\begin{table}[t!]
\centering
\caption{\textbf{Per-loop Crafter analysis} (Ouro-1.4B, unseen worlds). All rows evaluate final checkpoints with matched world sets; parentheses give mean gate-selected depth. Bold marks the highest achievement mean in each row. }
\label{tab:loop_probe}
\small
\renewcommand{\arraystretch}{1.08}
\setlength{\tabcolsep}{4.5pt}
\begin{tabular*}{\columnwidth}{@{\extracolsep{\fill}}lccc@{}}
\toprule
Exit & SFT & w/o Anchor & \ours{} \\
\midrule
Loop 1 & 0.00\sd{0.00} & \textbf{4.86}\sd{0.55} & 4.62\sd{0.47} \\
Loop 2 & 1.02\sd{0.21} & \textbf{6.74}\sd{0.71} & 5.08\sd{0.44} \\
Loop 3 & 6.10\sd{0.52} & \textbf{7.86}\sd{0.80} & 7.71\sd{0.58} \\
Loop 4 & 7.12\sd{0.60} & 6.31\sd{0.77} & \textbf{8.07}\sd{0.86} \\
\midrule
Gate & 7.21\sd{0.55} & 7.05\sd{0.86} & \textbf{8.24}\sd{0.73}  \\
\bottomrule
\end{tabular*}
\end{table}

\noindent\textbf{Per-loop analysis.}
\label{sec:depth_probe}
Improving shallow predictions does not automatically
improve every later loop or the performance obtained
through adaptive exits
(Table~\ref{tab:loop_probe}).
Without Anchor, distillation improves the first three
loops over SFT, but fourth-loop performance falls
from 7.12 to 6.31 achievements.
Moreover, gated play achieves 7.05, below the 7.86
achievements available at the third loop.
The improvement in intermediate predictions is therefore
not fully reflected in gated performance.

In contrast, \ours{} improves all four fixed-depth
readouts over SFT and achieves 8.24 achievements
with its learned exit policy, compared with 7.21 for SFT.
Its gated performance also exceeds its strongest
fixed-depth readout, which achieves 8.07.
These results illustrate why practice must address
both the predictions and their selection:
better shallow predictions alone are insufficient;
the policy must also learn when to use them.

\begin{table}[t]
\caption{\textbf{Repeated practice on the real robot.} The policy alternates between
10 practice episodes and one update, with no labels and no reward.
Starting policies and fixed evaluation scenarios are those of \cref{tab:robot_main}.}
\label{tab:continual}
\centering\footnotesize
\begin{tabular*}{\linewidth}{@{\extracolsep{\fill}}lcccc@{}}
\toprule
 & \multicolumn{2}{c}{Starting policy} & \multicolumn{2}{c}{Repeated practice}\\
\cmidrule(lr){2-3}\cmidrule(lr){4-5}
Method & Succ.\ (\%) & Iters & Succ.\ (\%) & Iters\\
\midrule
OPD    & 70.43{\scriptsize$\pm$1.90} & 5.50 & 64.95{\scriptsize$\pm$1.42} & 5.50\\
FLUENT & 74.57{\scriptsize$\pm$1.72} & 3.46 & 74.21{\scriptsize$\pm$0.97} & 3.21\\
\bottomrule
\end{tabular*}
\vspace{-3mm}
\end{table}

\vspace{1mm}
\noindent\textbf{Repeated practice.}
We test whether the gains of practice hold under repeated updates: ten cycles of 10 unlabeled practice episodes followed by one update, on scenarios disjoint from the evaluation scenarios (\cref{tab:continual}). \ours{} keeps its success rate (74.57\% to 74.21\%) while its executed iterations fall further (3.46 to 3.21). OPD loses 5.48 points (70.43\% to 64.95\%) with unchanged iterations. A single round of deep-to-shallow distillation is useful on its own: OPD raises success over SFT from 62.29\% to 70.43\% (\cref{tab:robot_main}), but this benefit does not persist under repeated updates. Anchor regularizes \ours{} toward the fixed initial model, and on a fixed context set the bound on this drift does not grow with the number of rounds, provided the objective does not exceed its initial value.

\section{Discussion}
\label{sec:discussion}

\noindent\textbf{The ruler moves with what it measures.}
Early-exit methods usually treat the last layer as a fixed target. In a looped model this fails: all loops share weights, so training the first loop also changes the final loop that serves as its ruler. Without Anchor, the model has a better prediction at its third loop, yet its gated play stays below it (7.05 vs.\ 7.86; \cref{tab:loop_probe}). Having a good prediction and choosing to use it are separate problems. Protecting the reference is not free: the more strongly it is protected, the less the first loop can adapt.

\vspace{1mm}
\noindent\textbf{The student is not a copy.}
A model that learns from itself, with no labels or rewards, ends up better than its teacher (fourth loop 8.07 vs.\ 7.12; \cref{tab:loop_probe}). A weight-shared model cannot memorize every situation separately, so it learns the teacher's judgment pooled over similar situations, which filters out the teacher's occasional mistakes, while Anchor largely preserves its confident decisions. Practice can correct the mistakes a teacher makes occasionally, but it cannot teach what the teacher does not know.

\vspace{1mm}
\noindent\textbf{Fluent is not the same as better.}
Practice lowered depth in every setting, but it did not always raise performance: Crafter and the real robot gained both, while LIBERO, which starts from about 92\% success, and the language models gained mainly depth. We hypothesize that two conditions matter: whether the initial model leaves room to improve, and whether the practiced situations resemble those at evaluation. Designing practice is then less a question of how much to practice than of what to practice.

\noindent\textbf{Limitations and future directions.}
The initial model provides a fixed teacher, so the quality of its predictions remains a constraint on the supervision available during practice.
 One way forward is to split practice into phases and refresh the teacher between them, while keeping it fixed within a phase so that the reference of the gate stays stable. Another is to give each loop a small number of its own parameters,
to explore whether this eases the trade-off between adapting the first loop and preserving the final loop.
Finally, extending the approach beyond looped models is a promising direction.
 Multi-exit transformers also share layers across exits~\citep{xin2020deebert,schuster2022confident},
so updates to shared layers can affect both early and final predictions, motivating joint study of prediction preservation and exit learning.

\section{Conclusion}
\label{sec:conclusion}

The same task is hard the first time and easy once it is familiar. We started from the view that difficulty depends not only on the input but also on experience. \ours{} is a self-distillation procedure that teaches looped models to produce useful predictions earlier and to decide when to use them. Translate uses the initial model's final-loop prediction as a frozen teacher for shallow predictions, Anchor regularizes the final-loop prediction toward the same teacher to stabilize the reference of Gate, and Gate adapts the exit policy to the updated predictions. No additional task labels or rewards are required. Across Crafter, LIBERO, real-robot manipulation, and language-model evaluation, \ours{} improves the performance--depth trade-off; on the real robot, this translates into higher success and shorter successful episodes. Practice can improve both predictions and the depth at which a model learns to use them. For models that repeatedly perform the same tasks, learning from experience to reduce computation is a practical requirement for efficient deployment. A robot that picks up the same cup for the hundredth time should not need to compute as long as it did the first time.

\section*{LLM Usage Statement}
We used OpenAI Codex and Anthropic Claude to translate the completed manuscript from a non-English language into English and to shorten sentences.
We also used these models to assist with the mathematical analysis: they helped formulate the statements and draft the proofs.
The authors examined each proof strategy, checked the results and references that the proofs rely on, and verified the correctness of every statement and proof.
The authors take full responsibility for the final manuscript, including its technical claims, proofs, references, experimental results, and interpretations.
\bibliography{main}
\bibliographystyle{ieeenat_fullname}

\end{document}